\documentclass{article}

\PassOptionsToPackage{square,numbers,sort&compress}{natbib}
\usepackage[position,preprint]{neurips_2026}

\usepackage[utf8]{inputenc}
\usepackage[T1]{fontenc}
\usepackage{hyperref}
\usepackage{url}
\usepackage{microtype}
\usepackage{graphicx}
\usepackage{booktabs}
\usepackage{flafter}
\usepackage{amsmath}
\usepackage{amssymb}
\usepackage{amsfonts}
\usepackage{mathtools}
\usepackage{amsthm}
\usepackage{multirow}
\usepackage[table]{xcolor}
\usepackage{xspace}
\usepackage{enumitem}
\usepackage{newunicodechar}
\usepackage[most]{tcolorbox}
\usepackage[capitalize,noabbrev]{cleveref}

\definecolor{rowgray}{gray}{0.85}
\definecolor{rowlightblue}{RGB}{215, 230, 255}
\definecolor{commentblue}{RGB}{30,100,200}

\newunicodechar{’}{'}

\definecolor{customblue}{HTML}{104862}

\newcounter{takeawaycounter}

\newenvironment{takeaway}[1][]{%
   \refstepcounter{takeawaycounter}%
   \begin{tcolorbox}[
       enhanced,
       breakable,
       title=#1,
       colback=white,
       colframe=customblue,
       colbacktitle=white,
       fonttitle=\bfseries,
       coltitle=black,
       attach boxed title to top left={yshift=-3mm, xshift=2mm},
       boxed title style={size=small, colback=white, frame hidden},
       sharp corners,
       rounded corners,
       arc=3mm,
       top=2mm,
       bottom=1mm,
       left=1mm,
       right=1mm,
       width=\linewidth,
   ]%
}{%
   \end{tcolorbox}
}

\theoremstyle{plain}

\theoremstyle{definition}

\theoremstyle{remark}

\title{Scalable Environments Drive Generalizable Agents}

\author{%
    \textbf{Jiayi Zhang}$^{1,2}$,
    \textbf{Fanqi Kong}$^{2,3}$,
    \textbf{Guibin Zhang}$^{4}$,
    \textbf{Maojia Song}$^{5}$,
    \textbf{Zhaoyang Yu}$^{2}$, \\
    \textbf{Jianhao Ruan}$^{1,2}$,
    \textbf{Jinyu Xiang}$^{1}$,
    \textbf{Bang Liu}$^{6,\dagger}$,
    \textbf{Chenglin Wu}$^{2,\dagger}$,
    \textbf{Yuyu Luo}$^{1,\dagger}$
    \\
    $^1$HKUST(GZ),
    $^2$DeepWisdom,
    $^3$PKU,
    $^4$NUS, \\
    $^5$SUTD,
    $^6$UdeM \& Mila
}

\begin{document}

\maketitle

\begin{abstract}
Generalizable agents should adapt to diverse tasks and unseen environments beyond their training distribution.
This position paper argues that such generalization requires environment scaling: expanding the distribution of executable rule-sets that agents interact with, rather than only increasing trajectories or tasks within fixed benchmarks.
Current scaling practices largely focus on collecting more experience or broader task sets under fixed interaction rules, leaving agents brittle when underlying interfaces, dynamics, observations, or feedback signals change.
The core challenge is therefore a world-level distribution shift: agents need systematic exposure to environments with meaningfully different executable rule-sets.
To clarify this challenge, we propose a unified taxonomy that separates trajectory scaling, task scaling, and environment scaling by their primary deliverables and by what changes in the executable rule-set.
Building on this taxonomy, we synthesize construction paradigms for scalable environments, contrasting programmatic generators that prioritize controllability and verifiability with generative world models that offer broader coverage and open-endedness.
We further outline how environment scaling can be coupled with stateful learning mechanisms, emphasizing learned update rules for cross-environment adaptation.
We conclude by discussing alternative perspectives and argue that scalable environments provide the essential substrate for measurable and controllable progress toward robust general agents.
\end{abstract}

\section{Introduction}
\label{sec:intro}

\begin{figure}[tbp]
    \centering
    \includegraphics[width=\linewidth,trim=30pt 0 14pt 0,clip]{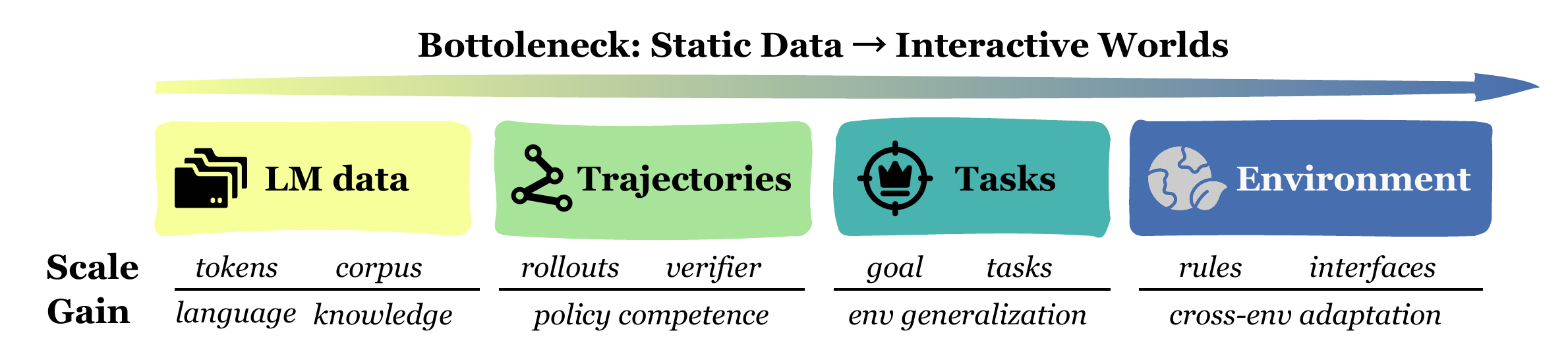}
    \caption{\textbf{Bottleneck shifts as we scale agents: from static data to interactive worlds.} Different scaling targets yield different primary gains: LM data improves language and knowledge; trajectories improve policy competence; tasks improve environment generalization within a fixed rule set; environments enable cross-environment adaptation under changing rules and interfaces.}
    \label{fig:bottleneck}
\end{figure}

Progress in general agents increasingly depends on acquiring diverse, high-quality experience through interaction, not only on scaling model parameters or static corpora~\cite{liu2025advances, silver2025welcome}.
Humans and artificial agents improve by acting in environments, observing consequences, and accumulating knowledge~\cite{gao2025survey, zhang2025memevolve, zhang2026harnessing, DBLP:journals/corr/abs-2505-07437}.
This shifts the bottleneck from model capacity to experience supply, namely, how to generate interaction data that reliably drives continual improvement.

A defining feature of general intelligence is competence across environments, transferring skills under changes in executable rules, including action interfaces, transition dynamics, and observation processes.
Humans exhibit this because our accessible environment distribution is effectively unbounded; we can enter and create new digital and physical worlds with different rules, tools, and interaction patterns~\cite{zhang2025autoenv}.
Crucially, humans do not merely accumulate experience; we actively explore and manipulate new environments to infer latent rules and affordances, building a highly adaptive world model~\cite{ying2025assessing} that supports fast skill acquisition and robust adaptation even in previously unseen worlds~\cite{richens2025general}.
In contrast, today's Language agents are typically trained and evaluated within a narrow set of fixed environments, and their performance often collapses when these rules shift.
This is not merely a shortage of data within one benchmark; it is a world distribution shift problem.

Recent work increasingly uses the term environment scaling to describe heterogeneous practices:
scaling trajectories in a fixed benchmark~\cite{li2025simulating, DBLP:conf/icml/Li0FXC0L25}, expanding task sets under fixed rules~\cite{sullivan2025procedural, DBLP:journals/pvldb/LiLCLT24}, or constructing new worlds with different rules altogether~\cite{song2026envscaler, dong2026agent, wu2026autowebworld, DBLP:journals/corr/abs-2510-23587}.
This conceptual overload obscures what is being scaled and makes comparisons difficult.
Trajectory scaling and task scaling are valuable and often necessary, but they mainly strengthen competence within a fixed rule set.
Without a systematic way to expand the environment distribution itself, the field risks overfitting to a few worlds while leaving cross-environment competence underspecified.
Scaling experience in a fixed world cannot substitute for scaling the world's agents must learn to live in.
Fig.~\ref{fig:bottleneck} summarizes this bottleneck shift by mapping each scaling target to its primary gain, from language and knowledge, to policy competence, to environment generalization, and ultimately to cross-environment adaptation as rules and interfaces vary.

Our position is that \textbf{scalable environments drive generalizable agents}: they provide a systematic way to generate, compose, and evaluate families of executable environments with meaningfully different rules. This is distinct from scaling trajectories or tasks within a fixed world. Rather than only improving competence under one rule set, environment scaling targets cross-world adaptation by exposing agents to changes in interfaces, dynamics, feedback, and social structure. As these factors vary, previously effective behaviors may fail, forcing agents to infer new rules and update their strategies through interaction. Scalable environments therefore make adaptation trainable and measurable through controlled rule shifts, while also supporting meta-learning, dynamics inference, and modular skill reuse across heterogeneous settings.

In this paper, we organize the landscape and outline a path forward.
First, we introduce an operational taxonomy that distinguishes \textbf{trajectory scaling,} \textbf{task scaling}, and \textbf{environment scaling} by their primary deliverables and what changes in the executable rule-set.
Second, we synthesize two major paradigms for scaling environments: \textbf{programmatic generators} that emphasize controllability and verifiability, and \textbf{generative world models} that offer broad coverage and open-endedness. And we discuss their tradeoffs and design principles.
Third, we clarify boundary cases where terminology diverges and summarize implications for learning under heterogeneity.
Our goal is to provide a unifying lens and a concrete call-to-action: environments should be treated as first-class artifacts that can be scaled, verified, and measured.

\paragraph{Contributions.}
\vspace{-0.55em}
\begin{itemize}[leftmargin=1.2em,itemsep=-0.25em,topsep=0.1em,parsep=0pt,partopsep=0pt]
  \item We argue that cross-environment competence is a prerequisite for general agents, and that scalable environments are therefore inevitable rather than optional.
  \item We propose an operational taxonomy that distinguishes trajectory, task, and environment scaling, including decision rules for boundary cases.
  \item We synthesize programmatic and generative paradigms for environment scaling and extract practical design principles and tradeoffs.
\end{itemize}
\vspace{-0.35em}

\begin{figure}[tbp]
    \centering
    \includegraphics[width=\linewidth,trim=24pt 7pt 24pt 8pt,clip]{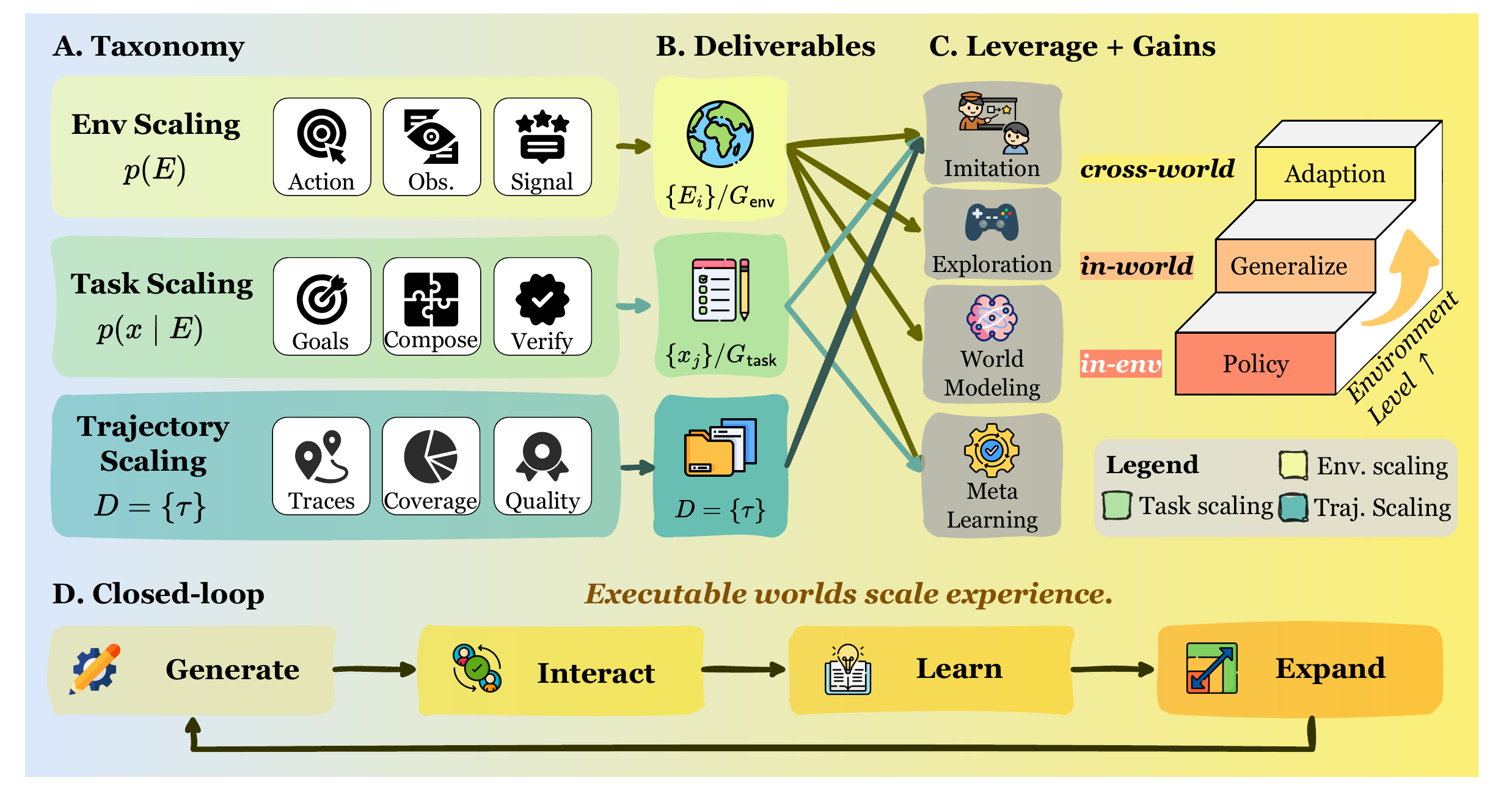}
    \caption{\textbf{Scalable environments as a substrate for scaling agents.}
We organize scaling into trajectory, task, and environment regimes (A) with corresponding deliverables (B), relate them to learning mechanisms and capability (C), and summarize the generate--interact--learn--scale loop (D).}
\label{fig:method_overview}
\end{figure}

\section{Preliminaries}
\label{sec:prelim}

We introduce a lightweight formulation that fixes notation for agent environment interaction. We model an \textsc{environment} as an \textbf{executable rule set} and an \textsc{agent} as a stateful system with a mental state that can be updated from experience. This view will serve as the basis for distinguishing trajectory, task, and environment scaling in Sec.~\ref{sec:taxonomy}.

\paragraph{Environment as an executable rule set.}
We model an environment $\mathcal{E}$ as an executable specification of interaction rules, which determines what actions are available, how the world evolves, what the agent observes, and what outcome signals the environment returns:
\begin{equation}
\mathcal{E} \triangleq \big(\mathcal{S},\; \mathcal{A},\; T,\; \mathcal{O},\; \Omega,\; \mathcal{Y}\big).
\label{eq:env_def}
\end{equation}
Here $\mathcal{S}$ is the state space; $\mathcal{A}$ the action interface defined by an interface contract such as tools, APIs, or commands; $T$ the transition dynamics; $\mathcal{O}$ the observation space with observation interface $\Omega$; and $\mathcal{Y}$ the \emph{environment signal} space. The environment signal may include rewards or scores, verifier outputs such as unit tests, termination flags, and structured execution outcomes. This definition emphasizes that an environment is not merely a collection of tasks, but a reproducible causal process that maps actions to transitions, observations, and outcome signals.

Although \(E\) is written as a construction-level executable object, the distinction relevant to agent learning is behavioral. 
Two implementations may use different hidden state spaces \(S\) or internal transition code \(T\), yet induce the same observable interaction process for the agent. 
Let \(h_t=(o_1,a_1,y_1,\ldots,o_t)\) denote the observable history before the next action. 
The rule-set \(E\) induces an agent-facing interaction kernel
\[
\mathcal{K}_E(o',y\mid h,a)
=
P(o_{t+1}=o',y_t=y\mid h_t=h,a_t=a;E).
\]
This kernel captures how observations and feedback signals evolve under executable actions. 
For environment scaling, a generated rule-set should alter this agent-facing process, such as the action interface, transition behavior, observation process, or feedback channel.

\paragraph{Tasks within an environment.}

A task $x$ specifies what the agent is asked to achieve in a fixed environment $\mathcal{E}$. We model tasks with an environment-conditioned distribution:
\[
x=(g,c,\rho_0,v),
\]
where \(g\) is the goal, \(c\) the constraints, \(\rho_0\) the initial conditions, and \(v\) the success verifier. Tasks are sampled as \(x\sim p(x\mid E)\), but do not alter the interaction kernel \(\mathcal{K}_E\). Thus, task scaling changes the objective, while environment scaling changes the interaction.

\paragraph{Trajectories as interaction traces.}
A trajectory $\tau$ is the sequence of interaction records produced when an agent runs in environment $\mathcal{E}$ on task $x$:
\begin{equation}
\tau \;=\; \big( (o_1, a_1, y_1),\; (o_2, a_2, y_2),\; \ldots \big),
\end{equation}
where $o_t \in \mathcal{O}$ is the observation, $a_t \in \mathcal{A}$ is an environment executable action under the interface contract, and $y_t \in \mathcal{Y}$ is the environment signal returned at step $t$. We use $\tau \sim p(\tau \mid \mathcal{E}, x, \pi)$ to denote the induced trajectory distribution under an agent policy $\pi$.

\subsection{Agents as Stateful Systems}
\label{sec:agent_formulation}

\paragraph{Agent state.}
An agent maintains a mental state $M_t \in \mathcal{M}$ at time $t$, capturing internal variables that mediate perception, decision making, and learning. 
We view $M_t$ as modular, \(M_t \;=\; \Big(M^{\mathrm{mem}}_t,\; M^{\mathrm{wm}}_t,\; M^{\mathrm{goal}}_t,\; M^{\mathrm{reward}}_t,\; \ldots\Big),\)
where modules may include memory, a world model, goals or motivation, and internal reward or feedback signals.

\paragraph{Policy and learning.}
We emphasize two external-facing capabilities of an agent: a \textsc{policy} for action selection and a \textsc{learning} mechanism that updates its internal state from experience~\cite{zhang2024aflow, ruan2026aorch, zhang2025memevolve,xiang2025self}. Concretely, interaction unfolds in discrete steps. Let $s_t \in \mathcal{S}$ denote the environment state. The environment provides an observation through its observation interface \(o_t = \Omega(s_t).\)
Given its current mental state, the agent selects an action that is directly executable under the environment interface, and the environment transitions according to its dynamics and returns an environment signal:
\begin{equation}
a_t \sim \pi(\cdot \mid M_t), \quad a_t \in \mathcal{A}, \qquad
s_{t+1} \sim T(s_t, a_t), \quad y_t \in \mathcal{Y}.
\label{eq:policy}
\end{equation}
Finally, the agent updates its mental state using the experience tuple, capturing learning such as memory writes, world model refinement, or parameter updates: \(M_{t+1} = L(M_t, a_t, o_{t+1}, y_t).\)
This formulation makes explicit that an agent is not only defined by action selection, but also by how it updates itself from interaction.

\section{The Current State of Environment Scaling: A Taxonomy}
\label{sec:taxonomy}

The term \emph{environment scaling} is increasingly overloaded, covering practices from generating traces in a fixed benchmark, to expanding task sets under fixed rules, to instantiating new rule-sets with heterogeneous interfaces and dynamics~\cite{huang2025environment}. This ambiguity makes comparisons difficult, since similar pipelines can target different axes of generalization. Building on Sec.~\ref{sec:prelim}, we categorize scaling by its \emph{primary deliverable}, namely the main object added to the learning loop (Fig.~\ref{fig:method_overview}).


Concretely, we distinguish three regimes.
\textbf{(i) Trajectory scaling} increases interaction traces and expands $p(\tau \mid \mathcal{E}, x)$ while keeping the underlying rule-set fixed.
\textbf{(ii) Task scaling} broadens the distribution of objectives within a fixed rule-set and expands $p(x \mid \mathcal{E})$.
\textbf{(iii) Environment scaling} expands the distribution of \emph{executable} rule-sets and expands $p(\mathcal{E})$, where $\mathcal{E}=(\mathcal{S},\mathcal{A},T,\mathcal{O},\Omega,\mathcal{Y})$.
In practice, these regimes can be composed, for example, new environments also yield new task families and traces once instantiated.
We therefore assign a method by what it primarily produces, namely additional traces $\tau$, a broadened task set $x \sim p(x\mid\mathcal{E})$, or new rule-sets $\mathcal{E}\sim p(\mathcal{E})$.
Ambiguous boundary cases, especially between task and environment scaling, are discussed separately in Sec.~\ref{sec:boundary-cases}.

\subsection{Trajectory Scaling}
\label{sec:traj-scaling}

\vspace{-0.25em}
\begin{takeaway}[Definition: Trajectory Scaling]
\textit{Trajectory scaling} expands or improves interaction traces $\tau$ under a largely fixed environment $\mathcal{E}$ and a fixed (or slowly varying) task family. Formally, it increases the coverage and quality of $p(\tau \mid \mathcal{E}, x)$ while keeping the executable rule-set $\mathcal{E}=(\mathcal{S},\mathcal{A},T,\mathcal{O},\Omega,\mathcal{Y})$ unchanged and without substantially altering $p(x\mid \mathcal{E})$.
\end{takeaway}
\vspace{-0.25em}

Under trajectory scaling, the main deliverable is \textbf{richer policy data}, including successful demonstrations, informative failures, and structured environment signals. These traces improve learning both directly, through imitation or reinforcement learning, and indirectly, by updating the agent's internal state through \(L\), such as memory, world models~\cite{silver2025welcome}, and internal evaluators. Together, these updates strengthen capability under a fixed rule set, without modifying \(\mathcal{A}\), \(T\), \(\Omega\), or \(\mathcal{Y}\).



\paragraph{Case studies.}
Trajectory scaling changes how traces are collected, filtered, or synthesized while keeping the executable rule-set fixed. One family uses execution-grounded collection, where agents act through the same interface \(A\) and environment signals \(Y\), such as success checks, unit tests, or external verifiers, are used to retain or rank useful traces~\cite{awadallah2025fara,he2025efficient,xu2026tooltext}. Another family uses simulation-based synthesis to cheaply expand attempts under the same dynamics \(T\), increasing the coverage of \(p(\tau\mid E,x)\) without introducing new rule-sets~\cite{chen2025scaling,li2025simulating}. In both cases, the primary artifact is richer trajectory data for improving \(\pi\) or updating the agent state through \(L\), rather than a new task or environment distribution.

\subsection{Task Scaling}
\label{sec:task-scaling}

\vspace{-0.25em}
\begin{takeaway}[Definition: Task Scaling]
\emph{Task scaling} expands the task distribution $p(x \mid \mathcal{E})$ under a fixed environment $\mathcal{E}$. Here, tasks vary in \textbf{goals, constraints, initial conditions}, while the executable rule-set remains unchanged, namely the action interface $\mathcal{A}$, transition dynamics $T$, observation space $\mathcal{O}$ and observation interface $\Omega$, and environment signals $\mathcal{Y}$ are fixed.
\end{takeaway}
\vspace{-0.25em}

By broadening the set of objectives within the same world, task scaling strengthens within-environment generalization: the agent learns to solve a wider range of tasks without altering the underlying rules. Task scaling also typically induces trajectory growth, since executing agents or demonstrators on new tasks naturally produces additional traces $\tau$.

\paragraph{Case Studies.}
Task scaling is commonly instantiated in two settings, depending on how tasks are represented and expanded under a fixed executable rule-set.
\textbf{(i) Open-ended environments.}
In tool-use and computer-use settings, the environment is open-ended in content while the interaction rules remain stable through a fixed interface \(\mathcal{A}\). Task scaling expands \(p(x\mid\mathcal{E})\) by generating new goals under the same interface constraints, often through task composition, seed-task evolution, or agent-proposed goals. Verification is usually heterogeneous, relying on partial checks or external procedures. A representative example is TaskCraft~\cite{shi2025taskcraft}, which generates tasks with controllable depth, width, and compositional structure under a fixed interaction interface. Related work expands or evaluates web and GUI task distributions under fixed interaction protocols~\cite{xie2025agentsynth, li2025websailor, sullivan2025procedural, deng2025interactcomp}.
\textbf{(ii) Closed environments.}
In embodied and gridworld-style settings, the executable rule-set \(\mathcal{E}\) is closed and fully specified, so tasks can be expressed as explicit goal specifications while keeping \(\mathcal{A}\), \(T\), \(\Omega\), and \(\mathcal{Y}\) fixed. Task scaling expands \(p(x\mid\mathcal{E})\) by procedurally instantiating structured goals, success predicates, or subgoal compositions within the same rules. A representative example is Beyond Fixed Tasks~\cite{furelos2025beyond}, which formalizes tasks as structured specifications under a fixed environment. Related work expands long-horizon activities and diverse success conditions in similar closed settings~\cite{jin2023mini}.

\subsection{Environment Scaling}
\label{sec:env-scaling}

\vspace{-0.25em}
\begin{takeaway}[Definition: Environment Scaling]
\emph{Environment scaling} expands the environment distribution $p(\mathcal{E})$ by generating or composing environments with \textbf{meaningfully different executable rule-sets}. Under this formulation, environment scaling corresponds to changes in one or more components of $\mathcal{E}=(\mathcal{S},\mathcal{A},T,\mathcal{O},\Omega,\mathcal{Y})$, such as the action interface $\mathcal{A}$, transition dynamics $T$, observation interface $\Omega$, or environment signals $\mathcal{Y}$.
\end{takeaway}
\vspace{-0.25em}

A change in the executable rule-set should also be meaningful from the agent's perspective. 
Let \(z_t=(o_t,a_t,y_t)\) denote one observable interaction record, including the observation, action, and environment signal. 
For a horizon \(H\) and a probing policy class \(\Pi\), we define
\[
d_{\mathrm{beh}}(E,E';H,\Pi)
=
\sup_{\pi\in\Pi}
D\left(
P(z_{1:H}\mid \pi,E),
P(z_{1:H}\mid \pi,E')
\right).
\]
If \(d_{\mathrm{beh}}(E,E';H,\Pi)=0\), then the two implementations are indistinguishable to the agent under the chosen probes, even if their hidden states or internal code differ. 
In this sense, environment scaling should expand behaviorally distinct interaction processes.

Task scaling varies goals, constraints, or initial conditions within a fixed agent-facing process. 
Environment scaling changes the process itself: the available actions, transition behavior, observation process, or feedback channel can shift, and the resulting environment should support repeated interactive rollouts. 
This distinction is what makes environment scaling directly relevant to cross-environment competence. 
As agents encounter behaviorally distinct rule-sets, a single fixed strategy is often insufficient; the agent must infer environment-specific regularities from interaction and update its internal state through the learning operator \(L\).


\paragraph{Case Studies.}
In contrast to task scaling, environment scaling expands $p(\mathcal{E})$ by introducing new executable rule-sets, and is therefore naturally upstream of both tasks and trajectories. When a rule-set $\mathcal{E}$ changes, it automatically leads to new task families $p(x \mid \mathcal{E})$ and new interaction traces $p(\tau \mid \mathcal{E}, x)$. We identify three common types of environment scaling based on which parts of $\mathcal{E} = (\mathcal{S}, \mathcal{A}, T, \mathcal{O}, \Omega, \mathcal{Y})$ are changed.

\textbf{(i) Interface Scaling.}
One route creates new tool-interactive environments by varying the action interface contract, including which executable actions are available and their constraints. EnvScaler~\cite{song2026envscaler} programmatically constructs new tool environments by modifying the tool interface and interaction rules, expanding $p(\mathcal{E})$ beyond goal-only variations. Related efforts similarly treat interface-level variation as a primary driver of environment diversity~\citep{froger2025scaling}.
\textbf{(ii) Rule-set Scaling.}
Another route scales environments by composing or generating new rule-sets that alter how the world evolves under interaction. AutoEnv~\cite{zhang2025autoenv} generates heterogeneous environments with distinct executable rules and measurement setups, making cross-environment generalization measurable and explicitly targeting environment-level shift rather than within-environment task variation. Related work in real-world, web, robotics, and co-evolution settings also induces new environment instances by modifying underlying rules and difficulty structures~\cite{dong2026agent, wu2026autowebworld, wang2023robogen, guo2025genenv}.
\textbf{(iii) Generative environment modeling.}
Learned generators and world models provide a generative pathway to environment scaling, where diversity is realized through learned simulators that produce varied executable dynamics and observations at scale~\cite{feng2025web,genie3}. Such methods offer broader coverage and open-endedness, but they require dedicated construction and evaluation protocols to ensure controllability and validity, which we discuss in Sec.~\ref{sec:how_and_eval}.

\subsection{Boundary Cases}
\label{sec:boundary-cases}

Many pipelines produce multiple artifacts at once. We resolve ambiguity with a two-stage rule. We first classify by the \emph{primary deliverable}, then use rule-set change as secondary evidence, checking whether the executable interaction rules for actions, transitions, observations, or signals change.

A common confusion is \textbf{trajectory scaling versus task scaling}. If the main deliverable is more or higher-quality trajectories under a largely fixed task family, we treat it as trajectory scaling. If the main deliverable is an expanded set of goals or specifications under a fixed rule-set, we treat it as task scaling, with traces induced downstream when the tasks are executed. Tool-use trace synthesis primarily delivers trajectories $\tau$~\cite{xu2026tooltext}, whereas task generators in web and GUI settings primarily deliver an expanded $p(x\mid\mathcal{E})$ and treat traces as secondary artifacts~\cite{shi2025taskcraft,li2025websailor,xie2025agentsynth}.

The other boundary is whether a method that generates tasks or traces actually creates a new environment. The key distinction is whether the output is a standalone, interactable environment instance that can be reused for repeated rollouts under a new rule-set, rather than a task-specific path or a collection of traces under a different action schema. If validation only ensures that the interface elements and execution paths needed for particular tasks work, it is task scaling. If the output is only trajectories, it is trajectory scaling. We treat a method as environment scaling only when it delivers a reusable environment instance with a modified interface or dynamics contract. Infinite Web~\cite{zhang2026infiniteweb} fits this boundary and is best categorized as task scaling, while tool-interactive environment synthesis methods that explicitly construct new environments fall under environment scaling~\cite{song2026envscaler}.

\section{Scalable Environments: Paradigms, Evaluation, and Usage}
\label{sec:how_and_eval}

\subsection{Programmatic Generators for Scalable Environments}
\label{sec:paradigm_programmatic}

Programmatic environment scaling expands the environment distribution by constructing executable rule-sets through code, rather than synthesizing additional trajectories within a fixed world. Concretely, a programmatic generator defines a mapping $\mathcal{E}=g_\theta(z)$, where $z\sim p(z)$ parameterizes layouts, entities, objectives, and interaction constraints, and the output $\mathcal{E}$ instantiates an executable interface and dynamics~\cite{verma2025measuring, zhang2025autoenv, song2026envscaler, wu2026autowebworld}. This style is particularly well-suited for environment scaling because it makes the rule-set itself explicit and manipulable: we can deterministically reproduce the same environment instance, apply targeted perturbations along specific axes of $\mathcal{E}$ (e.g., interface $A$, dynamics $T$, observation process $\Omega$, or signals $\mathcal{Y}$), and systematically generate controlled families of environments.

A key advantage of programmatic generators is \textbf{verifiability}. Since rules are encoded as executable programs, unit tests and invariants can validate transition consistency and the correctness of environment signals, including reward signals when applicable. This reduces noise caused by underspecified or unstable signals and enables rigorous stress tests under well-defined rule shifts. Moreover, programmatic generators provide \textbf{controllability} through modular parameterization and compositional operators, enabling calibrated curricula and factorized variation over environment axes rather than unconstrained diversity that is hard to diagnose.

A common concern is that purely programmatic worlds may lack semantic richness, limiting their ability to probe broad generalization. 
A useful abstraction is to decompose a programmatic environment into a verifiable rule scaffold $\bar{\mathcal{E}}$ and a content variable $c$:
\begin{equation}
\mathcal{E} = \textsc{Instantiate}(\bar{\mathcal{E}}, c).
\end{equation}
Here $\bar{\mathcal{E}}$ specifies the executable interaction contract (e.g., action interface and transition logic), while $c$ controls high-entropy semantics such as page contents, entities, and narratives. 
By embedding language-model calls inside the programmatic scaffold, we can sample
\begin{equation}
c \sim p_{\phi}\!\left(c \mid z, \bar{\mathcal{E}}\right)
\end{equation}
to generate diverse instances within the same underlying rule-set~\cite{feng2025web}. 
This augmentation diversifies semantic content in observations (and thus $p(x\mid\mathcal{E})$) while keeping $\bar{\mathcal{E}}$ fixed.
This hybridization increases semantic diversity without changing the core executable interaction, preserving reproducibility and verifiability while substantially enriching the environment distribution.

Beyond validity and control, programmatic generators expose a natural notion of \textbf{measurable environment complexity}. Environment scaling should not only increase the count of instances, but also the complexity of the underlying rule-sets being expanded. A useful lens is \emph{description complexity}: how succinctly an executable interaction contract can be specified. While Kolmogorov complexity $K(\mathcal{E})$ provides a conceptual ideal, it is not operational and is highly sensitive to language and library abstractions in real implementations. Therefore, we treat statistics such as lines of code (LoC) only as a coarse \emph{engineering footprint} of environment definitions, rather than a scientific complexity measure. In Sec.~\ref{sec:env_evaluation}, we complement such implementation-level statistics with \emph{behavioral} indicators (e.g., solution horizon, action branching, or feedback sparsity) that are less sensitive to coding style and dependencies. We report representative implementation footprints and use this perspective to motivate principled evaluation criteria in Sec.~\ref{sec:env_evaluation}.

\subsection{Generative World Models for Environment Scaling}
\label{sec:paradigm_generative}

Generative world models scale environments by producing explicit interactive worlds from compact specifications such as natural-language prompts. Formally, given a prompt $u$, the system samples an environment instance \(\mathcal{E} \sim p_{\phi}(\mathcal{E}\mid u).\)
In this sense, text-to-world systems~\cite{genie3} directly expand the support of the environment distribution $p(\mathcal{E})$ by instantiating new worlds on demand.

Crucially, the ability to generate explicit worlds often relies on learning how the world evolves under actions, analogous to physical laws in the physical laws in the real world. A common abstraction is a dynamics model that supports simulation or rollouts:
\begin{equation}
s_{t+1} \sim p_{\phi}(s_{t+1}\mid s_t, a_t), \qquad o_t \sim p_{\phi}(o_t\mid s_t),
\end{equation}
where approaches may differ in training signals and parameterizations, but all aim to capture transferable regularities of world evolution that enable reusable interaction.

This perspective highlights a practical prerequisite for generative environment scaling: learning such dynamics requires broad world data, which may come from videos and embodied logs, human-created interactive content, or synthetic corpora from programmatic simulators. In particular, programmatic environment generators can serve as controllable data engines that provide diverse, executable rollouts with known interfaces and feedback, while real-world video offers breadth and realism. Therefore, generative world models and programmatic generators are complementary: the former enables high-coverage novel worlds, while the latter supplies structured data and verification hooks that stabilize learning and evaluation. We return to unified evaluation criteria in Sec.~\ref{sec:env_evaluation}, focusing on validity, controllability, consistency, and verifiability of scaled environments.

\subsection{Evaluation Criteria for Scalable Environments}
\label{sec:env_evaluation}

Environment scaling expands a distribution over executable rule-sets $p(\mathcal{E})$, where $\mathcal{E}=(S,A,T,O,\Omega,Y)$ specifies the state space, action interface, transition dynamics, observation process, and feedback channel. 
Therefore, scalable environment suites should be evaluated not only by the number of generated instances, but by whether the induced $p(\mathcal{E})$ is usable and trustworthy for measuring cross-environment learning.
Appendix ~\ref{tab:env_loc} provides a minimal suite-level report illustrating practical statistics aligned with our evaluation criteria.
Concretely, we recommend reporting the following criteria.

\noindent\textbf{Executability.} A sampled $\mathcal{E}$ should instantiate a runnable interactive process with well-defined $(A,T,\Omega,Y)$ and support repeated rollouts. Key indicators include compilation/run success rate, crash-free interaction steps, and deterministic replay under fixed seeds and configurations.

\noindent\textbf{Signals.} Since the feedback channel $Y$ defines evaluation signals, unstable rewards collapse learning and measurement into noise. Suites should report \emph{when available} oracle coverage (fraction with checkable ground truth or constraints), reward reliability across repeated trials, and invariant violations that indicate inconsistencies in $T$ or $Y$.

\noindent\textbf{Coverage.} Environment scaling should expand the support of $p(\mathcal{E})$ along interpretable axes such as interface $A$, dynamics $T$, observation process $\Omega$, and feedback $Y$, rather than producing unconstrained diversity. Suites should also support controlled perturbations that vary one component while holding others fixed, enabling diagnostic stress tests under well-defined rule shifts.

\noindent\textbf{Complexity.} Since agents experience environments through trajectories $\tau \sim p(\tau\mid \mathcal{E},x,\pi)$, complexity should also be characterized behaviorally. We recommend reporting proxies such as minimal solution horizon, effective action branching, and feedback sparsity or delay (credit assignment length), which are less sensitive to implementation details than code-level statistics.

\noindent\textbf{Efficiency.} Scalable suites must be practical to expand $p(\mathcal{E})$ at scale. We recommend reporting generation throughput, cost per valid environment, and simulation cost per step, which determine sustainability of environment scaling.

Together, these criteria quantify environment scaling as a measurable knob over $p(\mathcal{E})$, capturing not only instance count but also validity, controllability, diversity, and signal quality, thereby enabling reliable study of general agents under world-level distribution shift.

\subsection{Building General Agents with Scalable Environments}
\label{sec:build_general_agents}
Scalable environments are not only larger testbeds, but also a training substrate for \emph{general agents}. This perspective is increasingly reflected in recent generalist agent systems, which scale training across many environments or games rather than a single fixed world, and evaluate transfer to unseen ones~\cite{bolton2025sima, magne2026nitrogen, zhang2026harnessing}. In our stateful formulation (Sec.~\ref{sec:agent_formulation}), an agent is defined by a policy conditioned on its internal state, $\pi(a_t \mid o_t, M_t)$, and an update operator $L$ that revises the internal state from experience. Environment scaling exposes the agent to world-level distribution shift $\mathcal{E}\sim p(\mathcal{E})$, turning general-agent training into optimizing performance \emph{in expectation over environments}, rather than within a single fixed world:
\begin{equation}
\max_{\pi,L}\; \mathbb{E}_{\mathcal{E}\sim p(\mathcal{E})}\Big[\; \mathbb{E}_{\tau \sim p(\tau \mid \mathcal{E},\pi,L)} \big[ \mathcal{Y}(\tau;\mathcal{E}) \big]\;\Big],
\end{equation}
where $\tau$ denotes an interaction trajectory and $\mathcal{Y}(\tau;\mathcal{E})$ is the environment-defined signal aggregated along the trajectory.

Even when the deployment environment is fixed, scalable generators can be used to construct a \emph{local neighborhood} of executable variants around a target environment $\mathcal{E}_0$. Let $p_{\text{loc}}(\mathcal{E}\mid \mathcal{E}_0)$ denote a distribution induced by controlled perturbations of layouts, entities, observation realizations, or other factors that preserve the core interaction contract. Training on this localized distribution provides abundant on-policy experience while keeping the task semantics anchored to the deployment domain:
\begin{equation}
\max_{\pi}\; \mathbb{E}_{\mathcal{E}\sim p_{\text{loc}}(\mathcal{E}\mid \mathcal{E}_0)}\Big[\; \mathbb{E}_{\tau \sim p(\tau \mid \mathcal{E},\pi)} \big[\mathcal{Y}(\tau;\mathcal{E}) \big]\;\Big].
\end{equation}
Compared to trajectory scaling inside a single \(\mathcal{E}_0\), this objective measures robustness under executable rule variations and reduces brittleness to deployment-time perturbations. 
Environment scaling also supplies adaptation experience: agents repeatedly encounter new rule sets and must update their internal state to remain effective. We capture this with the recursion,
\begin{equation}
M_{t+1} = L\!\left(M_t, o_t, a_t, \mathcal{Y}_t\right).
\qquad a_t \sim \pi(\cdot \mid o_t, M_t),
\end{equation}
Under world-level shift, the goal is not only to learn a fixed policy, but to learn an update mechanism with low post-shift regret across environments. A natural objective is \(\max_{\pi,L}\; \mathbb{E}_{E\sim p(E)}\!\left[J_E(\pi,L;K)\right],\)
where \(J_E(\pi,L;K)\) is the return after \(K\) adaptation steps via \(L\) in environment \(E\). This objective highlights the limitation of fixed-environment task scaling. For a policy \(\pi\) and update rule \(L\), let \(V_{\pi,L}^K(E)=\mathbb{E}_{x\sim p(x\mid E)}[J(E,x;\pi,L;K)]\). Task scaling around \(E_0\) optimizes \(V_{\pi,L}^K(E_0)\), whereas environment scaling optimizes \(J_{\mathrm{env}}(\pi,L)=\mathbb{E}_{E\sim p(E)}[V_{\pi,L}^K(E)]\). The signed gap can cancel across environments, so we measure rule-shift mismatch by
\[
\Delta_{\mathrm{shift}}(\pi,L;E_0)
=
\mathbb{E}_{E\sim p(E)}
\left[
\left|V_{\pi,L}^{K}(E)-V_{\pi,L}^{K}(E_0)\right|
\right].
\]
When \(p(E)\) includes distinct environments that affect return, this mismatch can be nonzero. Thus, scaling tasks inside \(E_0\) alone does not generally optimize cross-environment adaptation. Environment scaling addresses this gap with executable rule-set shifts.

\section{Alternative Views}
\label{sec:alternative-views}

A plausible alternative is that the limiting factor is not the scale of $p(\mathcal{E})$, but the quality of the environments and measurements we already have. Many successful suites are the product of careful human design, where progress comes from well-specified tasks, reliable signals, and strong instrumentation rather than from automated generation. From this perspective, effort is better spent on improving validity, reproducibility, and diagnosis of existing environments, since these properties directly determine whether evaluation is trustworthy and whether learning signals are stable. We agree that expert curation is often the fastest route to high-quality testbeds, and view scalable environment mechanisms as complementary: they matter primarily when we need low-marginal-cost expansion with controlled, reproducible rule shifts for systematic stress testing and measurement~\cite{huang2025environment,zhang2025autoenv}.

Even if scale is needed, another view is that scaling goals within a fixed world may deliver most practical benefit. When deployment follows a stable interaction contract, just expanding $p(x\mid\mathcal{E})$ can yield large gains by broadening goals, constraints, and difficulty. This matches the intuition that many agents do not need to generalize across arbitrary worlds, and that domain specialization can be both sufficient and more aligned with product needs and evaluation simplicity~\cite{shi2025taskcraft,li2025websailor,xie2025agentsynth}. We largely agree in fixed-domain settings. Our claim is narrower: when the target capability includes adaptation under rule-set shift, or when studying learning mechanisms that generalize beyond a single world, environment scaling becomes uniquely diagnostic and trainable.

A further alternative emphasizes grounding in real platforms and interaction data, arguing that synthetic environments risk overfitting to suite-specific artifacts and may not transfer. In this view, the most valuable experience comes from in-the-wild interfaces, observations, and signals, and scalable environments should mainly serve as scaffolds for collecting or structuring such data rather than as the primary object to scale. We see this as compatible with our position. Programmatic generators provide controllable and verifiable rule scaffolds, while learned world models and content generation can enrich interaction experience at scale, including by embedding model calls inside executable scaffolds~\cite{feng2025web,genie3}. Even if the bottleneck lies in agent-side adaptation and learning machinery, developing and validating such mechanisms still benefits from repeated, controlled rule-set variation that uncontrolled streams rarely provide~\cite{song2026envscaler,silver2025welcome,DBLP:conf/acl/HongLLLWZLCZWZZ25}.

\section{Conclusion}
We argued that scaling agents toward generalization requires scaling \emph{environments}, not only data, trajectories, or within-world tasks. By defining an environment as an executable rule set and separating trajectory scaling, task scaling, and environment scaling by their deliverables and rule changes, we clarify common confusions and provide practical criteria for classification and reporting. We further outlined two complementary paths for expanding environment distributions, programmatic generators, and generative world models, and proposed a suite-level checklist of executability, signals, coverage, complexity, and efficiency to make progress comparable. We hope this perspective helps re-center the field on scalable, controllable, and verifiable environments as a foundation for building agents that adapt under changing rules, interfaces, and feedback.

\bibliographystyle{plainnat}
\bibliography{cited}

\newpage
\appendix

\section{Appendix}

\begin{table}[htbp]
\centering
\caption{\textbf{Suite-level reporting statistics for representative scalable environment suites.}
We summarize validity and reproducibility (Executable, Consistency), interface footprint (\#Actions), signal and observation types (Reward, Observation), LLM-in-Env usage, and implementation footprint (Language and average lines of code, LoC).
$\bullet$/$\circ$ denote yes/no for binary properties (Executable, Consistency, LLM).
``--'' indicates statistics that are not applicable or not reported in the original suite description.
LoC is reported as a coarse engineering footprint and should not be interpreted as a scientific complexity measure.}
\label{tab:env_loc}
\resizebox{\linewidth}{!}{
\begin{tabular}{lcccccccc}
\toprule
\textbf{Method}
& \textbf{Executable}
& \textbf{Consistency}
& \textbf{\#Actions}
& \textbf{Reward}
& \textbf{Observation}
& \textbf{LLM}
& \textbf{Language}
& \textbf{LoC} \\
\midrule
GG-Bench
& $\bullet$ & $\bullet$ & -- & Binary & Text & $\circ$ & Python & 126.6 \\
EnvScaler
& $\bullet$ & $\bullet$ & 18.6 & Binary & Text & $\circ$ & Python & 862.8 \\
AutoEnv
& $\bullet$ & $\bullet$ & 6.1 & Binary \& Dense & Text+2D+3D & $\circ$ & Python & 973.1 \\
Web World Model
& $\bullet$ & $\bullet$ & -- & None & Text & $\bullet$ & HTML/JS & 4556.6 \\
Genie3
& $\bullet$ & $\circ$ & -- & None & 3D & -- & -- & -- \\
\bottomrule
\end{tabular}}
\end{table}

\begin{table}[htbp]
\centering
\caption{Scaling Methods Comparison. We categorize existing methods by their scaling level: Trajectory (Traj.), Task, or Environment (Env.). Scaling Level is multi-select as some methods scale multiple levels simultaneously. Scaling Objects refers to the primary artifacts being expanded. Target Capability indicates the agent capability that each method aims to improve.}
\label{tab:scaling-methods}
\begin{tabular}{lcccl}
\toprule
\multirow{2}{*}{Method} 
& \multicolumn{3}{c}{Scaling Level} 
& \multirow{2}{*}{Scaling Objects}  \\
\cmidrule(lr){2-4}
& Traj. & Task & Env. &  \\
\midrule
Simulating Env~\cite{li2025simulating} & $\bullet$ & $\circ$ & $\circ$ & Simulated feedback \\
DreamGym~\cite{chen2025scaling} & $\bullet$ & $\bullet$ & $\circ$ & Curriculum tasks \\
Fara-7B~\cite{awadallah2025fara} & $\bullet$ & $\bullet$ & $\circ$ & Verified traces  \\
Unlocking Implicit~\cite{xu2026tooltext} & $\bullet$ & $\circ$ & $\circ$ & Tool definitions  \\
AutoBencher~\cite{li2024autobencher} & $\circ$ & $\bullet$ & $\circ$ & Evaluation instances  \\
Beyond Fixed Tasks~\cite{furelos2025beyond} & $\circ$ & $\bullet$ & $\circ$ & Goal specifications  \\
TaskCraft~\cite{shi2025taskcraft} & $\bullet$ & $\bullet$ & $\circ$ & Multi-hop tasks \\
Procedural Env Gen~\cite{sullivan2025procedural} & $\circ$ & $\bullet$ & $\circ$ & Tool configurations  \\
Mini-BEHAVIOR~\cite{jin2023mini} & $\circ$ & $\bullet$ & $\circ$ & Activity compositions \\
Infinite Web~\cite{zhang2026infiniteweb} & $\bullet$ & $\bullet$ & $\circ$ & Web pages \\
InteractComp~\cite{deng2025interactcomp} & $\circ$ & $\bullet$ & $\circ$ & Ambiguous queries \\
AgentSynth~\cite{xie2025agentsynth} & $\bullet$ & $\bullet$ & $\circ$ & GUI tasks  \\
AgentGen~\cite{hu2025agentgen} & $\bullet$ & $\bullet$ & $\bullet$ & Env-task pairs \\
EnvScaler~\cite{song2026envscaler} & $\bullet$ & $\bullet$ & $\bullet$ & Tool interfaces  \\
AutoEnv~\cite{zhang2025autoenv} & $\bullet$ & $\bullet$ & $\bullet$ & Rule distributions  \\
Agent-World~\cite{dong2026agent} & $\bullet$ & $\bullet$ & $\bullet$ & Synthesized environments \\
AutoWebWorld~\cite{wu2026autowebworld} & $\bullet$ & $\bullet$ & $\bullet$ & Web FSMs \\
Web World Models~\cite{feng2025web} & $\circ$ & $\circ$ & $\bullet$ & World states \\
Genie 3~\cite{genie3} & $\circ$ & $\circ$ & $\bullet$ & Interactive worlds  \\
ARE~\cite{froger2025scaling} & $\bullet$ & $\bullet$ & $\bullet$ & Env abstractions \\
RoboGen~\cite{wang2023robogen} & $\bullet$ & $\bullet$ & $\bullet$ & Scenes, supervisions \\
\bottomrule
\end{tabular}
\end{table}

\end{document}